\UseRawInputEncoding
\pdfoutput=1

\documentclass[11pt]{article}

\usepackage[preprint]{acl}

\usepackage{times}
\usepackage{latexsym}

\usepackage[T1]{fontenc}

\usepackage[utf8]{inputenc}

\usepackage{microtype}

\usepackage{inconsolata}

\usepackage{graphicx}
\usepackage{multirow}
\usepackage{booktabs}
\usepackage{url}
\usepackage[utf8]{inputenc}
\usepackage{caption}
\usepackage{amsmath}
\usepackage{amsthm}
\usepackage{algorithm}
\usepackage{algorithmic}
\usepackage[switch]{lineno}
\usepackage{inconsolata}
\usepackage{microtype}
\usepackage[T1]{fontenc}
\usepackage{subfigure}
\usepackage{diagbox}
\usepackage{color}
\usepackage{xspace}
\usepackage{listings}
\usepackage{dsfont}
\usepackage{float}
\usepackage{appendix}
\usepackage{algorithm}
\usepackage{algorithmic}

\usepackage{arydshln}
\usepackage{booktabs,amsfonts,dcolumn}
\usepackage{multirow}
\title{\textit{re}CSE: Portable Reshaping Features for Sentence Embedding in Self-supervised Contrastive Learning}

\author{Fufangchen Zhao\textsuperscript{\rm 1, \rm 2}, Jian Gao\textsuperscript{\rm 1, \rm 2}, Danfeng Yan\textsuperscript{\rm 1, \rm 2} \\
  \textsuperscript{\rm 1} Beijing University of Posts and Telecommunications,  Beijing, China \\
  \textsuperscript{\rm 2} State Key Laboratory of Networking and Switching Technology \\
  \texttt{zhaofufangchen@bupt.edu.cn} \\}

\begin{document}
\maketitle
\begin{abstract}
We propose \textit{re}CSE, a self supervised contrastive learning sentence representation framework based on feature reshaping. This framework is different from the current advanced models that use discrete data augmentation methods, but instead reshapes the input features of the original sentence, aggregates the global information of each token in the sentence, and alleviates the common problems of representation polarity and GPU memory consumption linear increase in current advanced models.  In addition, our \textit{re}CSE has achieved competitive performance in semantic similarity tasks. And the experiment proves that our proposed feature reshaping method has strong universality, which can be transplanted to other self supervised contrastive learning frameworks and enhance their representation ability, even achieving state-of-the-art performance. \footnote{Our code is available at  \url{https://github.com/heavenhellchen/reCSE}} 
\end{abstract}

\section{Introduction}

Self-supervised sentence representation tasks \citep{le2020contrastive}, which involve obtaining vector embeddings with rich semantic information from raw text in a self-supervised manner and can adapt to various downstream tasks without fine-tuning, have gained renewed attention due to the rise of contrastive learning \citep{chopra2005learning,hadsell2006dimensionality,oord2018representation}. Previous studies have directly employed pre-trained language models (PLM), such as BERT \citep{devlin2018bert} and RoBERTa \citep{liu2019roberta}, to derive high-quality sentence representations which still perform poorly in specific downstream tasks (e.g. semantic similarity task \citep{reimers2019sentence}) without fine-tuning. Consequently, contrastive learning facilitate the emergence of more advanced methods \citep{gao2021simcse,yan2021consert} which subsequently assume a dominant position in the domain of sentence representation tasks.

The concept of contrastive learning suggests that the crux of self-supervised sentence representations learning hinges on the acquisition of suitable positive and negative samples from unlabeled data. \citet{yan2021consert} employs various surface-level data augmentation techniques to derive positive samples from the original sentences. In contrast, \citet{gao2021simcse} adopts a more sophisticated approach that implicitly treats dropout \citep{hinton2012improving} as the baseline method for data augmentation. Specifically, \citet{gao2021simcse} feeds each sentence from a batch into a pre-trained BERT or RoBERTa model twice, applying independently sampled dropout masks for each pass. Consequently, \citet{gao2021simcse} considers the two distinct embeddings derived from the same original sentence as "positive sample pairs". Meanwhile, sentences from the same mini-batch that are not part of these pairs are categorized as "negative samples".
\begin{figure}[t]
   \centering
   \includegraphics[width=\linewidth]{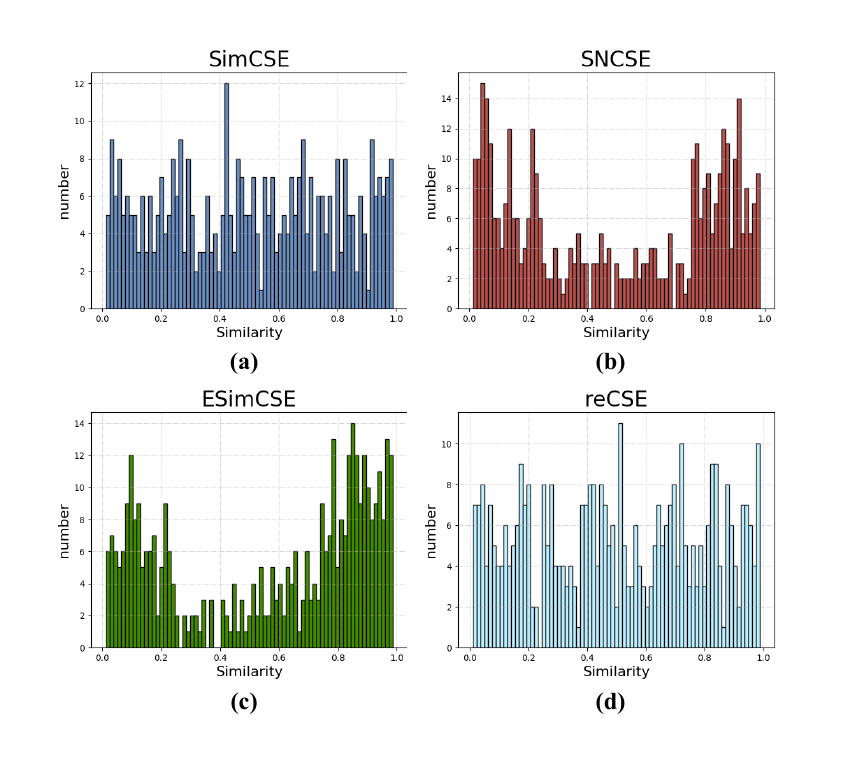}
    \caption{The distribution of representation polarity test results. The distribution of the framework (b, c) based on discrete data augmentation shows polarity (concavity), and the distribution of the basic SimCSE and our \textit{re}CSE (a, d) is relatively uniform.}  
    \label{motivation_1}
    \vspace{-5pt}
\end{figure}

\begin{figure}[ht]
   \centering
   \includegraphics[width=\linewidth]{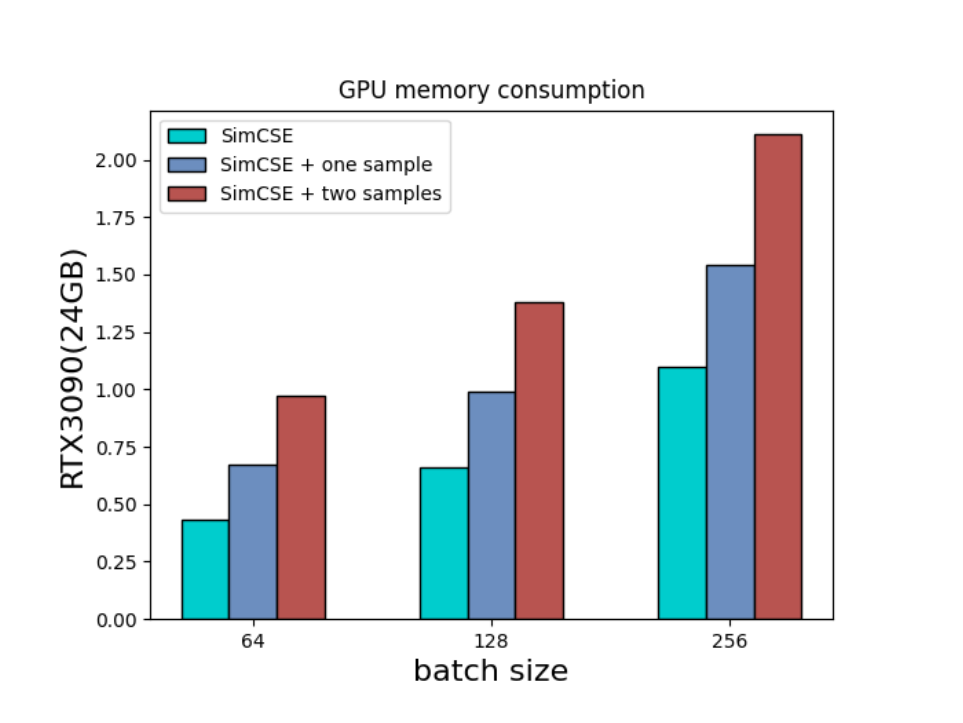}
    \caption{The impact of discrete data augmentation on GPU memory consumption. The y-axis scale is measured in RTX3090 (24GB) units. As more types of additional samples are introduced, the GPU memory consumption for training also increases linearly.}
    \label{motivation_2}
    \vspace{-5pt}
\end{figure}

Further researchers \citep{wang2023sncse,wu2021esimcse,shi2023osscse,chuang2022diffcse,xu2024blendcse,zhao2024simct,zhuo2023whitenedcse} have identified several drawbacks associated with the generation of positive samples based on the same original sentence and dropout. These insights have spurred the development of numerous advanced works that address these issues, resulting in significant performance improvements. Although these works target different problems, they share a \textit{discrete augmentation} method: the introduction of supplementary samples tailored to the specific challenges they aim to overcome. Although the approach of incorporating supplementary samples has yielded impressive and intuitive performance in downstream tasks,  it concurrently introduces novel challenges:
\begin{itemize}
    \item The incorporation of supplementary samples is anticipated to augment the polarity of the sentence representation model. In essence, from the perspective of semantic similarity, models augmented with supplementary samples are more likely to assign higher or lower scores to sentences based on their similarity. To verify this issue, we employ GPT-4 \citep{achiam2023gpt} to generate 400 sentence pairs, each labeled with a similarity score ranging from 0 to 3, where 0 indicates distinct semantics and 3 signifies identical semantics. Subsequently, we utilized SimCSE \citep{gao2021simcse}, SNCSE \citep{wang2023sncse} and ESimCSE \citep{wu2021esimcse}, where SNCSE and ESimCSE incorporates additional samples based on SimCSE, to evaluate these pairs. The final statistical results are depicted in Figure \ref{motivation_1} (a,b,c),  which shows that after introducing supplementary samples, the predicted results have obvious polarity. 
    \item Furthermore, introducing supplementary samples is anticipated to escalate the GPU memory requirements for model training. Specifically, this augmentation increases the number of sentences processed in each training batch, consequently amplifying the GPU memory overhead. Notably, the diversity of supplementary samples is positively correlated with the extend of the increased memory overhead. We further conduct a preliminary experiment to substantiate our hypothesis. Employing SimCSE as the foundational model, we simulate the incorporation of supplementary samples through the application of additional dropout layers to the input sentences. We monitor the GPU memory consumption throughout the model training process. The results of this experiment are delineated in Figure \ref{motivation_2}.
\end{itemize}
To address the aforementioned challenges, we consider the following issues:
\begin{itemize}
    \item Is there a strategy that can enhance the understanding of the overall semantics of sentences by contrastive learning frameworks, thereby enhancing their capacity for sentence representation without the incorporation of supplementary samples?
    \item How to design the contrastive learning framework that enhances sentence representation without resulting in a linear increase in GPU memory.
\end{itemize}

To sum up, we introduce \textit{re}CSE, a novel contrastive learning framework for sentence embedding that eschews the introduction of supplementary samples in favor of portable feature \textbf{re}shaping. Specifically, a single word is insufficient to encapsulate the full semantic content of a sentence. Merely tokenizing and encoding a sentence does not effectively capture its global information. Consequently, our proposed \textit{re}CSE approach reconstructs the tokenized features and integrates the comprehensive information of the sentence into the features. We integrate this process with a contrastive learning loss to fortify the original contrastive learning framework, thereby enhancing its capacity for sentence representation. In addition, to mitigate GPU memory consumption, we design the feature reshaping process as an independent module, or "pendant", of the original contrastive learning framework. This design operates independently, without necessitating additional inputs to the models embedded within the original framework. Through this "time for space" methodology, we effectively compress the GPU memory consumption throughout the training process.

In short, we make the following contributions:
\begin{itemize}
    \item We propose \textit{re}CSE, a contrastive learning framework that enhances sentence representation capabilities through feature reshaping alone, eliminating the need for supplementary samples.
    \item We innovatively decouple the feature reshaping process from the contrastive learning framework, enabling the feature reshaping and embedding models to operate independently, which significantly reduces GPU memory consumption.
    \item We are surprised to discover that feature reshaping exhibits portability, and our discovery corroborate through experimental validation on alternative sentence representation models. 
\end{itemize}

\section{Related Work and Background}

\subsection{Sentence Representation Learning}
As a foundational task in the domain of natural language processing (NLP), sentence representation learning has garnered sustained interest over time. \citet{wu2010semantics} and \citet{tsai2012bag} employ a bag-of-word model to represent sentences, whereas \citet{kiros2015skip} and \citet{hill2016learning} categorize the task directly as a context prediction challenge. Recently, the proliferation of pre-trained language models \citep{devlin2018bert,liu2019roberta,brown2020language} has led many researchers to opt for utilizing models such as BERT \citep{devlin2018bert} and RoBERTa \citep{liu2019roberta} to derive sentence representations. Although sentence representations generated by these pre-trained models are theoretically versatile for adaptation to any downstream task, achieving competitive performance without subsequent fine-tuning remains challenging. Furthermore, some studies \citep{ethayarajh2019contextual,yan2021consert} have identified that employing the [CLS] token directly as the sentence representation or utilizing average pooling in the last layer may result in \textbf{anisotropy}, a phenomenon where the learned embeddings converge into a limited region. Subsequently, methods such as BERT-Flow \citep{li2020sentence} and BERT-Whitening \citep{su2021whitening} have been proposed and have effectively mitigated this issue. Most recently,  contrastive learning \citep{gao2021simcse} has emerged as the predominant approach for sentence representation tasks.

\subsection{Contrastive Learning for NLP}
Contrastive learning \citep{he2020momentum,chen2020simple} has garnered significant success in natural language processing by minimizing the distance between positive samples and maximizing the separation between negative samples \citep{gao2021simcse,wu2021esimcse,yan2021consert,kim2021self}. Recently, the concepts of \textbf{alignment} and \textbf{uniformity} \citep{wang2020understanding} have been introduced as metrics for assessing the quality of representations derived from contrastive learning.  Alignment evaluates the proximity of positive sample pairs, whereas uniformity assesses the impact of anisotropy on the spatial distribution of embeddings.

Based on the aforementioned research, SimCSE \citep{gao2021simcse} is introduced by researchers as a seminal contribution to the field. It leverages the inherent randomness of the dropout noise to enrich the latent space of semantically aligned sentences, thereby generating diverse sentence representations that constitute positive pairs. For an exhaustive treatment of this topic, Please refer to section \ref{simcse}. Subsequent researches have further enhanced the quality of sentence representations based on SimCSE. ESimCSE \citep{wu2021esimcse} mitigates model bias induced by sentence length by incorporating supplementary samples through word repetition. Concurrently, SNCSE \citep{wang2023sncse} bolsters the capability of sentence representation by introducing the negation of original sentences as supplementary negative samples. OssCSE \citep{shi2023osscse} directly introduces two different supplementary samples to counteract the bias stemming from the uniformity of surface structures, \textbf{etc}. These methods uniformly leverage the introduction of supplementary samples to bolster sentence representation capability. While the impact is considerable, this strategy may escalate computational demands and introduce polarity within sentence representations. On the other hand, PromptBERT \citep{jiang2022promptbert} enhances the quality of sentence embeddings produced by BERT \citep{devlin2018bert} in SimcSE framework, employing a prompt-based method \citep{zhou2022debiased}. DCLR \citep{zhou2022debiased} concentrates on refining the capacity for negative sample selection. ArcCSE \citep{zhang2022contrastive} directly targets the optimization of the objective function. Nonetheless, these methodologies are predominantly contingent upon the inherent quality of the training data, exhibit challenges in portability, and demonstrate only modest enhancements in the capability of sentence representation.
\subsection{Background: SimCSE}
\label{simcse}
This section provides a detailed introduction to the foundational framework employed in our study: SimCSE \citep{gao2021simcse}.

In the context of two semantically similar sentences, we define these as a sentence pair, denoted as $\left\{x_i, x_i^+\right\}$, and consider this pair to constitute a positive sample. The central tenet of SimCSE involves utilizing identical sentence to forge positive samples pair,  i.e., $x_i=x_i^+$.  Note that there is a dropout mask placed on the fully-connected layers and the attention probabilities in Transformer. Consequently, the essence of constructing positive samples in SimCSE lies in encoding the same sentence $x_i$ twice with distinct dropout masks $z_i$ and $z_i^+$, thereby yielding two distinct embeddings that form a positive sample pair:
\begin{equation}
    \mathbf{h}_{i}=f_{\theta}\left(x_{i},z_i\right), \mathbf{h}_{i}^{+}=f_{\theta}\left(x_{i}, z_i^{+}\right)
\end{equation}
For each sentence within a mini-batch of size $N$, the contrastive learning objective with respect to $x_i$, given the embeddings $\mathrm{h}_{i}$ and $\mathrm{h}_{i}^+$, is formulated as follows:
\begin{equation}
    \ell_{i}=-\log\frac{e^{\mathrm{sim}\left(\mathbf{h}_{i}, \mathbf{h}_{i}^{+}\right)/\tau}}{\sum_{j=1}^{N} e^{\mathrm{sim}\left(\mathbf{h}_{i}, \mathbf{h}_{j}^{+}\right)/\tau}}
\end{equation}
where $\tau$ is a temperature hyperparameter and $\mathrm{sim}\left(\mathbf{h}_{i}, \mathbf{h}_{i}^{+}\right)$ is the similarity metric, which is typically the cosine similarity function\footnote{$\operatorname{sim}\left(\mathbf{h}_i, \mathbf{h}_i^{+}\right)=\frac{\mathbf{h}_i^{\top} \mathbf{h}_i^{+}}{\left\|\mathbf{h}_i\right\| \cdot\left\|\mathbf{h}_i^{+}\right\|}$}.

\section{Method}

\begin{figure*}[t]
   \centering
   \includegraphics[width=\linewidth]{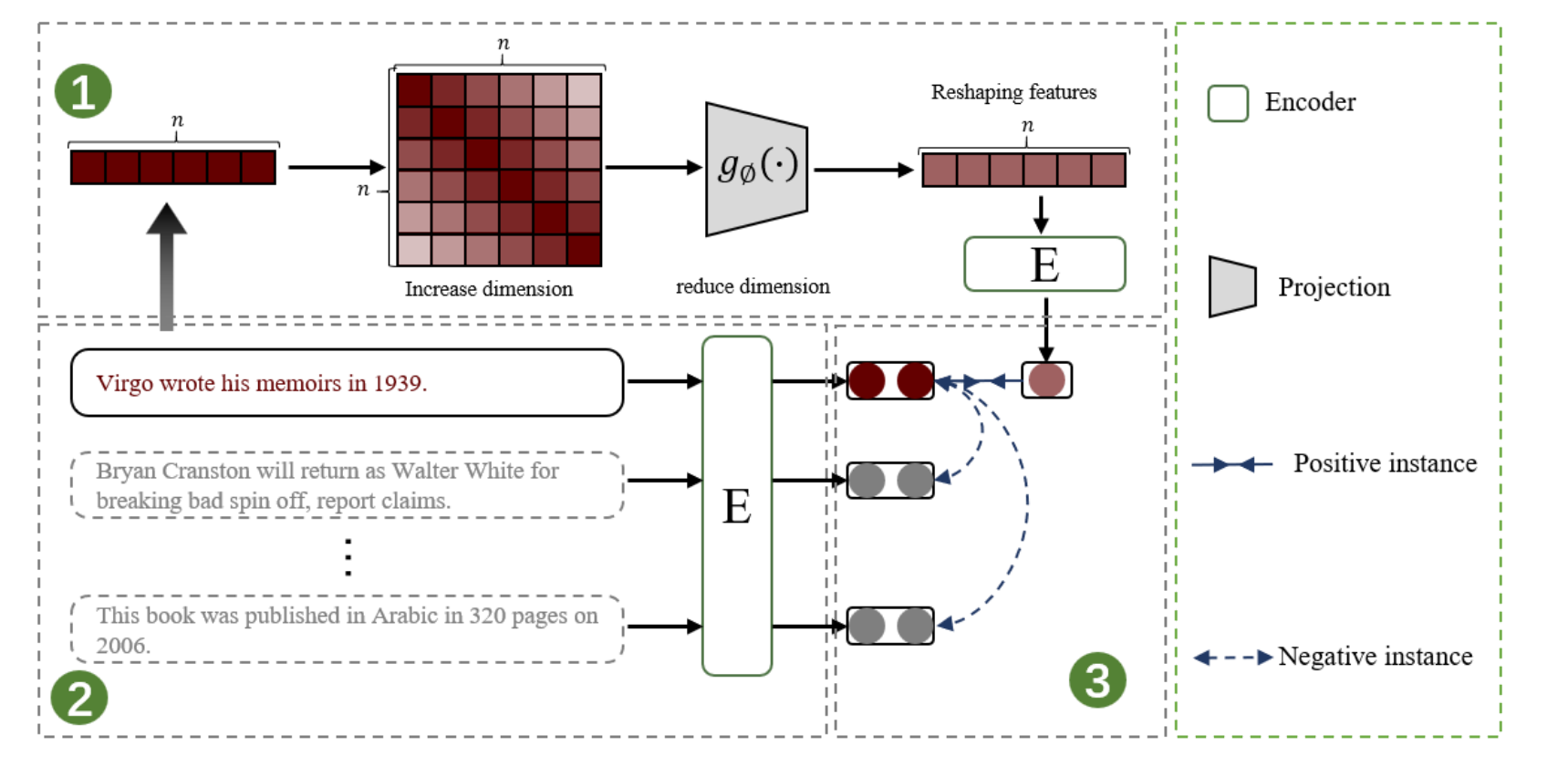}
    \caption{The main framework of \textit{re}CSE. We adopt a modular design to reduce GPU memory consumption}
    \label{method}
\end{figure*}

The \textit{re}CSE framework we proposed is shown in Figure \ref{method}, which can be divided into three components: (1) the feature reshaping, (2) the dropout-based original data augmentation, (3) and the contrastive learning mechanism that integrates these reshaped features. Next we will provide a detailed introduction to each component.

\subsection{Feature reshaping}
\begin{figure}[t]
   \centering
   \includegraphics[width=\linewidth]{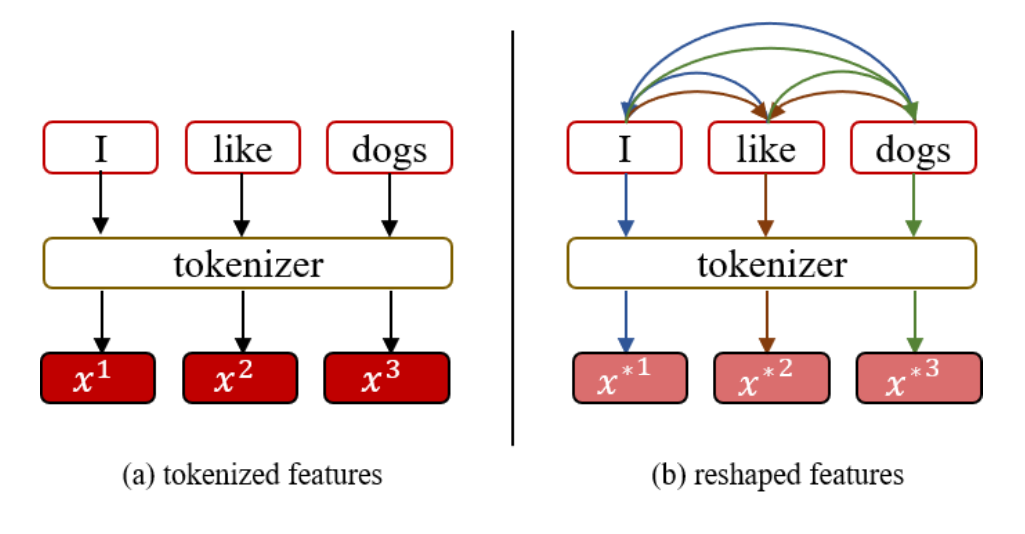}
    \caption{The original input features are based solely on a single token (a), while the reshaped features contain the global information of each token in the sentence (b).}  
    \label{feature_reshaping}
    \vspace{-5pt}
\end{figure}
The feature reshaping part is the main innovation of our proposed \textit{re}CSE. This part aims to reshaping the input features without generating supplementary samples, thereby enhance the focus on global information contained within the sentences, as shown in Figure \ref{feature_reshaping}. The ultimate objective is to integrate these reshaped features into contrastive learning framework.

Next, we offer an exhaustive exposition on our feature reshaping:

Given a sentence of $n$ tokens $s=\{s^{1},s^{2},\dots,s^{n}\}$, we initially employ a tokenizer to derive the original feature $x = \{x^{1},x^{2},\dots,x^{n}\}$:
\begin{equation}
    \{x^{i}\}_{i=1}^n = \mathrm{Tokenizer}(\{s^{i}\}_{i=1}^n\})
\end{equation}
Note that the original feature $x$ can be conceptualized as an $1 \times n$ matrix, where each element represents  information pertaining to the respective token. Based on this conceptualization, we enhance the feature by augmenting its dimensionality and densifying it, thereby transforming $x$ into an $n \times n$ matrix $X$:

\begin{equation}
\label{dimensionality}
    X = \sqrt{x^{\mathrm{T}} \cdot x}
\end{equation}
The matrix $X$ derived from Eq. 4\label{dimensionality} can be partitioned into two components:
\begin{equation}
    X = \operatorname{diag}(x^{1},x^{2},\cdots,x^{n}) + \hat{X}
\end{equation}
where the first part is a diagonal matrix, with its diagonal elements corresponding to those of the original features of corresponding tokens $\{x^{i}\}_{i=1}^n$, and the second part is a symmetric matrix $\hat{X}$ with diagonal elements of $0$, with its off-diagonal elements representing the correlations features between the original features of distinct tokens $(x^{i},x^{j})_{i \neq j}^{n}$. In essence, the $i$-th column of matrix $X$ comprises the features of token $x^i$ as well as the correlation features with respect to other tokens.

Ultimately, we need to compress matrix $X$ back to its original dimensions, thereby extracting the reshaped features, which are input into the encoder:
\begin{equation}
    x^* = g_{\phi}(X)
\end{equation}
where $g$ represents a linear projection operation applied per column of the matrix, designed to aggregate the $n \times n$ matrix $X$ along its columns via a linear transformation, thereby reduce it to original $1 \times n$ dimension.

\subsection{Dropout-based Data Augmentation}
In the processing of the original sentence, We first use different prompts \citep{brown2020language} to enhance sentence representation:
\begin{equation}
\centering
\begin{split}
The\;sentence\;:\;"\;s\;"\;mean\;[MASK].\\
The\;sentence\;of\;"\;s^+\;"\;means\;[MASK]. \\
\end{split}
\end{equation}
We adhere to the SimCSE methodology \citep{gao2021simcse} and employ dropout as the minimum data augmentation unit. For a set  of sentence features $\{x_i\}_{i=1}^N$, we construct positive sample pair by encoding  each input sentence feature $x_i$ twice with different dropout masks:
\begin{equation}
\boldsymbol{h}^{z}=f_\theta(x_i, z), \boldsymbol{h}^{z^{\prime}}=f_\theta(x_i, z^{\prime})
\end{equation}
where $z$ and $z^{\prime}$ denote different dropout masks, $f_{\theta}(\cdot)$ is a pre-trained language encoder such as BERT and RoBERTa. To sum up, distinct embeddings $\boldsymbol{h}^{z}$ and $\boldsymbol{h}^{z^{\prime}}$ based on identical input, constitute the positive sample pair we need. Because of the adoption of prompts, we ultimately embed the hidden state of the special $\mathrm{[MASK]}$ token $h_{[\textit{MASK}]}$ as the input sample. We added an additional MLP layer with tanh activation function on $h_{[\textit{MASK}]}$ to obtain $h$: 
\begin{equation}
\begin{split}
\boldsymbol{h}^{z}=Tanh(MLP(\boldsymbol{h}^{z}_{[\textit{MASK}]}))\\  \boldsymbol{h}^{z^{\prime}}=Tanh(MLP( \boldsymbol{h}^{z^{\prime}}_{[\textit{MASK}]}))
\end{split}
\end{equation}
Additionally, other inputs in the same mini-batch are categorized as negative samples.
\subsection{Contrastive Learning with Reshaped Features}
We employ the infoNCE loss \citep{he2020momentum} as the training objective for our proposed \textit{re}CSE, which embodies the concept of contrastive learning through a straightforward cross-entropy loss. Furthermore, for a collection of input sentences $\{s_i\}_{i=1}^N$, the procedures delineated in the preceding sections facilitate the acquisition of three distinct embedding representations: the original sentence embedding $h_i^{z}$, the positive sample embedding $h_i^{z^{\prime}}$, and the reshaped embedding $h_i^{*}$. 

For $h_i^{z}$ and $h_i^{z^{\prime}}$, our training objectives are as follows:
\begin{equation}
\ell_{\mathrm{CL}}=-\log \frac{e^{\operatorname{cos\_sim}\left(h_i^{z}, \boldsymbol{h}_i^{z^{\prime}}\right) / \tau}}{\sum_{j=1}^N e^{\operatorname{cos\_sim}\left(\boldsymbol{h}_i^{z}, h_j^{z^{\prime}}\right) / \tau}}
\end{equation}
where $N$ is the size of the mini-batch, $\tau$ is a temperature parameter. 

For the reshaped embedding representation $h_i^*$, we consider it as an additional positive sample and endeavor to approximate it to $h_i^{z}$ and $h_i^{z^{\prime}}$ as closely as possible. The specific equation is as follows:
\begin{equation}
\ell_\mathrm{\textit{re}}=-\sum_{Z \in\left\{z, z^{\prime}\right\}} \log \frac{e^{\operatorname{cos\_sim}\left(h_i^{Z}, h_i^*\right) / \tau^{\prime}}}{\sum_{j=1}^N e^{\operatorname{cos\_sim}\left(h_i^{Z}, h_j^*\right) / \tau^{\prime}}}
\end{equation}
where $\tau^{\prime}$ is a temperature parameter. Let $\lambda$ denote the trade-off hyperparameter that balances the two objectives, we formulate the final loss as:
\begin{equation}
    \ell = \lambda\ell_{\mathrm{CL}} + (1-\lambda) \mathrm{max}(\ell_{\mathrm{CL}}, \ell_\mathrm{\textit{re}})
\end{equation}
We don't mandate $\ell_\mathrm{\textit{re}}$ as an obligatory training objective, instead, we prioritize its optimization during the initial stages of training. Specifically, when the $\ell_\mathrm{\textit{re}}$ value is minimal, our optimization strategy focus solely on $\ell_{\mathrm{CL}}$.


\section{Experiments}
\subsection{Setup}
Following \citet{gao2021simcse}, we randomly select 1,000,000 sentences from English Wikipedia to form our input sentence corpus. We conduct experiments utilizing BERT \citep{devlin2018bert} and RoBERTa \citep{liu2019roberta} as encoders, respectively. During evaluation phase, we employ semantic similarity tasks (STS) to evaluate the sentence representation capability of the proposed \textit{re}CSE framework. The STS task is designed to measure the semantic similarity between sentences and represents one of the most widely utilized benchmark datasets for assessing self-supervised sentence embeddings. The task encompasses seven sub-tasks, specifically STS-12 to STS-16, STS-B, and SICK-R \citep{agirre2012semeval,agirre2013sem,agirre2014semeval,agirre2015semeval,agirre2016semeval,cer2017semeval,marelli2014sick}. Our evaluation method is as follows:
\begin{equation}
\rho = 1 - \frac {\sum\limits_{i=1}^{n} {(R_E^i-R_G^i)}^2} {\frac {1} {6} (n^3-n)}
\end{equation}
where $R_E^i$ represents the estimated rank for each sentence pair, while $R_G^i$ denotes the ground-truth rank. The metric $\rho$ which ranges from -1.0 to 1.0, indicates the sophistication of the sentence embedding and the semantic comprehension ability of the model; a higher value suggests superior performance. We compute $\rho$ for each subtask individually and the determine the average of $\rho$ as the primary indicator.

In detail, our \textit{re}CSE is implemented through Torch 1.7.1 and huggingface transformers \citep{wolf-etal-2020-transformers}. We conduct experiments using batch sizes of 128 and smaller on computing nodes equipped with Nvidia GTX 3090 GPUs, while other experiments are performed on a single A100 GPU. Additionally, for all experiments, we set the dropout rate to 0.1. We train the model for a total of 3 epochs, evaluate every 250 steps, and select the model parameters that demonstrated the highest performance.

\subsection{Main Results}
\begin{table*}[t!]
    \begin{center}
    \centering
    \small
\begin{tabular}{lcccccccc}
    \toprule

        \multicolumn{9}{c}{\it{Semantic Textual Similarity (STS) Benchmark}}\\
         \midrule
         \textbf{Model-BERT$_{base}$} & \textbf{STS12} & \textbf{STS13} & \textbf{STS14} & \textbf{STS15} & \textbf{STS16} & \textbf{STS-B} & \textbf{SICK-R} & \textbf{Avg.} \\
   
    \midrule
         SimCSE \citep{gao2021simcse} & 68.69 &	82.05 &	72.91 &	81.15 &	79.39 &	77.93 &	70.93 &	76.15 \\
        
        MoCoSE \citep{cao2022exploring} & {71.58} & 81.40 & 74.47 & {{83.45}} & 78.99 & 78.68 & {72.44} & 77.27 \\

        InforMin-CL \citep{chen2022informin-cl} & 70.22 & {{83.48}} & {75.51} & 81.72 & {79.88} & 79.27 & 71.03 & 77.30 \\

        MixCSE \citep{zhang2022unsupervised} & 71.71 & 83.14 & 75.49 & 83.64 & 79.00 & 78.48 & 72.19 & 77.66\\
DCLR$^*$ \citep{zhou2022debiased} & 70.81 & 83.73 & 75.11 & 82.56 & 78.44 & 78.31 & 71.59 & 77.22\\

ArcCSE$^*$ \citep{zhang2022contrastive}& 72.08 & 84.27 & 76.25 & 82.32 & 79.54 & 79.92 & 72.39 & 78.11 \\
        
PCL$^*$ \citep{WuPCL22} & {72.74} & {83.36} & {76.05} & {83.07} & 79.26 & {79.72} & {72.75} & 78.14 \\

ESimCSE$^*$ \citep{wu2021esimcse} & \textbf{73.40} & 83.27 & {77.25} & 82.66 & 78.81 & 80.17 & 72.30 & {78.27} \\
DiffCSE$^*$ \citep{chuang2022diffcse}& {72.28} & {84.43} & {76.47} & {83.90} & {80.54} & {80.59} & 71.29 & 78.49 \\
miCSE \citep{klein2022micse} & {71.71} &	{83.09} &	{75.46} &	{83.13} &	{80.22} &	{79.70} &	{73.62} &	{78.13} \\
PromptBERT \citep{jiang2022promptbert} & {71.56} & {84.58} & {76.98} & {84.47} & \textbf{80.60} & \textbf
{81.60} & {69.87} & {78.54}\\
SNCSE$^*$ \citep{wang2023sncse} & {70.67} & \textbf{84.68} & {76.99} & {83.69} & {80.51} & {81.35} & {74.77} & {78.97}\\
OssCSE$^{\dag,*}$ \citep{shi2023osscse} & {71.78} & {84.40} & \textbf{77.71} & \textbf{83.95} & {79.92} & {80.57} & \textbf{75.25} & \textbf{79.08}\\

(ours) \textit{re}CSE & {72.03} &	{84.61} &	{75.46} &	{83.72} &	{80.45} &	{80.71} &	{74.61} &	{78.68} \\
\midrule
\textbf{Model-RoBERTa$_{base}$} & \textbf{STS12} & \textbf{STS13} & \textbf{STS14} & \textbf{STS15} & \textbf{STS16} & \textbf{STS-B} & \textbf{SICK-R} & \textbf{Avg.} \\
\midrule
SimCSE \citep{gao2021simcse} & {70.16} & {81.77} & {73.24} & {81.36} & {80.65} & {80.22} & {68.56} & {76.57}\\
DCLR$^*$ \citep{zhou2022debiased} & {70.01} & {83.08} & {75.09} & {83.66} & {81.06} & {81.86} & {70.33} & {77.87} \\
PCL$^*$ \citep{WuPCL22} & {71.13} & {82.38} & {75.40} & {83.07} & {81.98} & {81.63} & {69.72} & {77.90}\\
ESimCSE$^*$ \citep{wu2021esimcse} & {69.90} & {82.50} & {74.68} & {83.19} & {80.30} & {80.99} & {70.54} & {77.44} \\
DiffCSE$^*$ \citep{chuang2022diffcse} & {70.05} & {83.43} & {75.49} & {82.81} & {82.12} & {82.38} & {71.19} & {78.21}\\
PromptRoBERTa \citep{jiang2022promptbert} & \textbf{73.94} & {84.74} & {77.28} & \textbf{84.99} & {81.74} & {81.88} & {69.50} & {79.15} \\
SNCSE$^*$ \citep{wang2023sncse} & {70.62} & {84.42} & {77.24} & {84.85} & {81.49} & {83.07} & {72.92} & {79.23} \\
OssCSE$^{\dag,*}$ \citep{shi2023osscse} & {72.28} & \textbf{85.27} & \textbf{79.51} & {84.77} & \textbf{82.32} & \textbf{83.55} & \textbf{75.54} & \textbf{80.46} \\
(ours) \textit{re}CSE & {73.72} &	{84.11} &	{76.47} &	{83.97} &	{81.42} &	{83.14} &	{72.52} &	{79.19} \\
\bottomrule
    \end{tabular}
\end{center}
\caption{The results of semantic similarity task test. $*$ denotes the use of \textit{discrete augmentation}, and $\dag$ means this framework isn't open-source.}
    \label{main_results}
\end{table*}
Our main experiment results are presented in Table \ref{main_results}. It is observable that, with the exception of OssCSE \citep{shi2023osscse}, which is not open source, our proposed \textit{re}CSE exhibits a marginally lower average performance compared to state-of-the-art models, such as SNCSE \citep{wang2023sncse}, yet it retains strong competitiveness. However, it is important to note that the models currently exhibiting the highest average performance, such as SNCSE and OssCSE, employ discrete data augmentation techniques. These techniques introduce additional samples to bolster the sentence representation capabilities. And a distinct polarity in the ability to represent sentences is evident. Upon excluding these models that utilize discrete augmentation (marked with $*$ in Table \ref{main_results}) our proposed \textit{re}CSE demonstrates the most superior performance and there is almost no introduction of polarity in representational ability as well (We will prove this in the next section). Furthermore, our method, which does not introduce any additional samples, is theoretically universal and can be integrated with any form of discrete enhancement, which is the "portability" mentioned in this work. We have also substantiated this claim through experiments detailed in next section. 

Specifically, our proposed \textit{re}CSE has demonstrated significant improvements over the baseline model SimCSE in all subtasks. When employing BERT as the encoder, our method realizes the maximum enhancement on the SICK-R benchmark, outperforming SimCSE by 3.68. Even though the STS-16 benchmark shows the least improvement, it still reflects a 1.06 gain. Utilizing RoBERTa as the encoder, our method garners a 3.96 improvement on the SICK-R benchmark and a 0.77 improvement on the STS-16 benchmark. These results substantiate the effectiveness of our approach.

\section{Further Discussion}
\subsection{Sentence Representation Polarity Analysis}
In the preceding introduction, we identify that numerous sentence representation models exhibit polarity. Specifically, from the perspective of semantic similarity, these models consistently assign either higher or lower scores to a sentence pair. We posit that the underlying cause of this polarity is attributable to the presence of discrete data augmentation methods. We have also demonstrated this in the preceding introduction. Furthermore, since our \textit{re}CSE framework does not employ discrete data augmentation, it is theoretically devoid of the introduction of polarity.

To substantiate our claim, we conduct additional test using our \textit{re}CSE on 400 sentences generated by GPT-4 (mentioned in the Introduction section). The result is depicted in Figure \ref{motivation_1} (d), which clearly demonstrate that our proposed \textit{re}CSE framework does not introduce polarity, thereby highlighting the superiority of our approach.

\subsection{Reducing GPU Memory Consumption Analysis}
\begin{figure}[t]
   \centering
   \includegraphics[width=\linewidth]{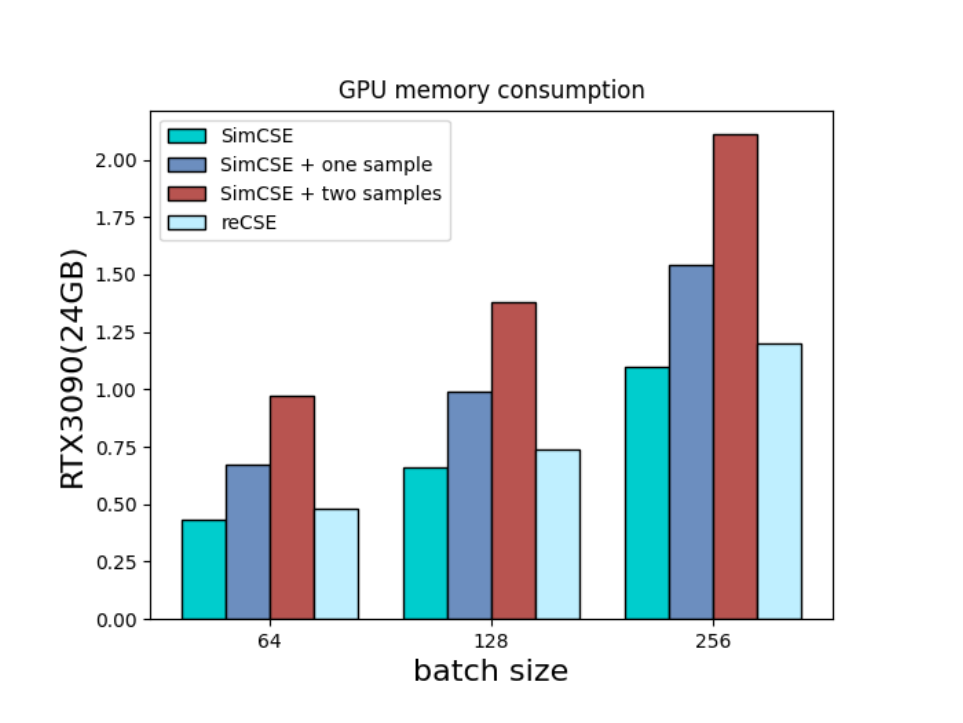}
    \caption{Test results of \textit{re}CSE on GPU memory consumption.}  
    \label{GPU_consumption}
    \vspace{-5pt}
\end{figure}

\begin{table*}[t]
    \begin{center}
    \centering
    \small
\begin{tabular}{lcccccccc}
    \toprule

        \multicolumn{9}{c}{\it{Semantic Textual Similarity (STS) Benchmark}}\\
         \midrule
         \textbf{Model-BERT$_{base}$} & \textbf{STS12} & \textbf{STS13} & \textbf{STS14} & \textbf{STS15} & \textbf{STS16} & \textbf{STS-B} & \textbf{SICK-R} & \textbf{Avg.} \\
         \midrule
         \midrule
         ESimCSE \citep{wu2021esimcse} & {73.40} & 83.27 & {77.25} & 82.66 & 78.81 & 80.17 & 72.30 & {78.27} \\
         ESimCSE + \textit{Feature reshaping} & \textbf{73.67} & \textbf{84.09} & \textbf{78.28} & \textbf{83.76} & \textbf{78.89} & \textbf{81.25} & \textbf{73.34} & \textbf{79.03}\\
         \midrule
         SNCSE \citep{wang2023sncse} & \textbf{70.67} & {84.68} & {76.99} & {83.69} & \textbf{80.51} & {81.35} & \textbf{74.77} & {78.97}\\
         SNCSE + \textit{Feature reshaping} & {70.54} & \textbf{84.99} & \textbf{77.84} & \textbf{84.71} & {80.49} & \textbf{81.87} & {73.64} & \textbf{79.26} \\
   
     \midrule
\textbf{Model-RoBERTa$_{base}$} & \textbf{STS12} & \textbf{STS13} & \textbf{STS14} & \textbf{STS15} & \textbf{STS16} & \textbf{STS-B} & \textbf{SICK-R} & \textbf{Avg.} \\
   \midrule
      \midrule
ESimCSE \citep{wu2021esimcse} & {69.90} & {82.50} & {74.68} & {83.19} & {80.30} & \textbf{80.99} & {70.54} & {77.44}\\
ESimCSE + \textit{Feature reshaping} & \textbf{70.11} & \textbf{82.92} & \textbf{75.77} & \textbf{84.07} & \textbf{80.71} & {80.98} & \textbf{71.34} & \textbf{78.37}\\
\midrule
SNCSE   \citep{wang2023sncse} & {70.62} & {84.42} & {77.24} & {84.85} & \textbf{81.49} & {83.07} & {72.92} & {79.23} \\
SNCSE + \textit{Feature reshaping} & \textbf{70.63} & \textbf{85.04} & \textbf{78.25} & \textbf{84.86} & {81.25} & \textbf{83.98} & \textbf{73.07} & \textbf{79.88}\\

\bottomrule
    \end{tabular}
\end{center}
\caption{Portability test results based on semantic similarity task.}
    \label{portability_test}
\end{table*}
In our analysis of SimCSE, we identify that the encoder based on pre-trained language models, such as BERT or RoBERTa, is the primary contributor to GPU memory consumption during training. Specifically, when the batch size is set to $N$, each encoder input comprises $2 \times N$ sentences. Advanced works, such as SNCSE, OssCSE, etc., often employ discrete data augmentation to introduce supplementary samples, which increase the encoder's input during training to $3 \times N$ even $4 \times N$ sentences or higher. This approach increases the consumption of GPU memory and loses generality, which is not conducive to the continued research of subsequent researchers.

Based on the above considerations, we decompose the \textit{re}CSE framework into three sequential steps, as depicted in Figure 3. To curtail GPU memory consumption during training, we took inspiration from the CoCo dataset's processing methodology \citep{lin2014microsoft}, opting to isolate the initial step entirely. Initially, we reshaped and stored the input sentence features. Subsequently, we augmented the original sentences with dropout. Ultimately, the third step involved extracting and training the stored reshaped representations. Since the two encoders within the \textit{re}CSE framework do not operate concurrently, an increase in the required training time is inevitable. Nevertheless, the GPU memory consumption will not surpass the maximum cost in first two steps. Theoretically this design has led to significant decrease in GPU memory consumption compared to prior studies and provide higher reference for subsequent researches.

To substantiate our argument, we execute a comparative GPU memory consumption test on the proposed \textit{re}CSE with batch size $[64,128,256]$ as in the Introduction section, and the final result is shown in Figure \ref{GPU_consumption}. It can be seen that our \textit{re}CSE has almost the same memory consumption as SimCSE \citep{gao2021simcse}, which demonstrates the effectiveness of our method in reducing GPU memory consumption.

\subsection{Portability analysis}
Through further analysis, we determine that our proposed feature reshaping method is universally applicable and is compatible with commonly used discrete data augmentation. To validate this hypothesis, we transplant our feature reshaping method into several state-of-the-art contrastive learning models that are currently available as open-source, including  ESimCSE \citep{wu2021esimcse} and SNCSE \citep{wang2023sncse}. Subsequently,and conduct semantic similarity testing. The test results are shown in Table \ref{portability_test}. It is evident that both ESimCSE and SNCSE exhibit significant performance enhancements following the introduction of the proposed feature reshaping. Notably, when employing BERT as the encoder, SNCSE's performance even surpasses that of the state-of-the-art model (OssCSE \citep{shi2023osscse}, 79.08) despite not being open source.

\section{Conclusion}
In this paper, we investigate the challenges associated with current advanced self-supervised contrastive learning frameworks for sentence representation. Specifically, we introduce two key issues: the polarity in representational capacity due to discrete augmentation and the linear escalation of GPU memory consumption during training as a result of incorporating additional samples. To address these challenges, we propose a novel feature reshaping and introduce \textit{re}CSE, a self-supervised contrastive learning framework for sentence representation based on feature reshaping. Experimental results demonstrate that our proposed \textit{re}CSE framework achieves competitive performance in semantic similarity tasks without a corresponding increase in GPU memory consumption. Additionally, we have showcased the portability of our feature reshaping method across other self-supervised contrastive learning frameworks. In general, we anticipate that our research will facilitate future researchers' recognition of the limitations inherent in discrete data augmentation methods, thereby inspiring the development of more universally applicable and efficacious approaches.
\section{Limitations}
Our work employs a "time for space" to mitigate GPU memory consumption. However, this approach results in an extended training duration. Furthermore, the projection algorithm in the feature reshaping is currently rudimentary, indicating scope for further refinement and optimization.

\bibliography{reCSE}




\end{document}